\documentclass[10pt,twocolumn,letterpaper]{article}

\usepackage{cvpr}
\usepackage{times}
\usepackage{epsfig}
\usepackage{graphicx}
\usepackage{amsmath}
\usepackage{amssymb}

\usepackage{multirow}
\usepackage{comment}
\usepackage{caption}
\usepackage{subcaption}
\usepackage{multirow}
\usepackage{microtype}

\usepackage{wrapfig}

\newcommand{\p}[0]{\phi}

\usepackage{color}

\definecolor{dkmag}{rgb}{0.5,0,0.5}

\newcommand{\francisco}[1]{#1}

\usepackage[pagebackref=true,breaklinks=true,letterpaper=true,colorlinks,bookmarks=false]{hyperref}

\cvprfinalcopy 

\def\cvprPaperID{1528} 

\ifcvprfinal\fi
\begin{document}

\title{Deep Exemplar 2D-3D Detection by Adapting from Real to Rendered Views}

\author{Francisco Massa \textsuperscript{1}
\quad\quad
Bryan C. Russell \textsuperscript{2}
\quad\quad
Mathieu Aubry\textsuperscript{1,}\textsuperscript{3} \vspace{4mm}\\
\textsuperscript{1} 
\'Ecole des Ponts ParisTech \thanks{ Universit\'{e} Paris-Est, LIGM (UMR CNRS 8049), ENPC, F-77455 Marne-la-Vallée. This work was carried out in IMAGINE, a joint research project between Ecole des Ponts ParisTech (ENPC) and the Centre Scientifique et Technique du B\^atiment (CSTB).} \quad\quad
\textsuperscript{2} Adobe Research \quad\quad
\textsuperscript{3} UC Berkeley\\
}

\maketitle

\begin{abstract}
This paper presents an end-to-end convolutional neural network (CNN) for 2D-3D exemplar detection.
We demonstrate that the ability to adapt the features of natural images to better align with those of CAD rendered views is critical to the success of our technique.
We show that the adaptation can be learned by compositing rendered views of textured object models on natural images.
Our approach can be naturally incorporated into a CNN detection pipeline and extends the accuracy and speed benefits from recent advances in deep learning to 2D-3D exemplar detection.
We applied our method to two tasks: instance detection, where we evaluated on the IKEA dataset~\cite{Lim13}, and object category detection, where we out-perform Aubry \etal~\cite{Aubry14b} for ``chair'' detection on a subset of the Pascal VOC dataset.
\end{abstract}

\section{Introduction}

Recently, Aubry \etal~\cite{Aubry14b} performed object category detection by exemplar alignment with a large library of 3D object models. The 
aligned models often approximately matched the style of the depicted objects and allowed 3D information, such as hidden object surfaces and object pose, to be propagated to the 2D images. Such a result is useful for 3D scene reasoning and may potentially be used in applications such as object manipulation in robotics and model-based object image editing in computer graphics~\cite{Kholgade}.

%


\begin{figure*}
\center
\includegraphics[width=\linewidth]{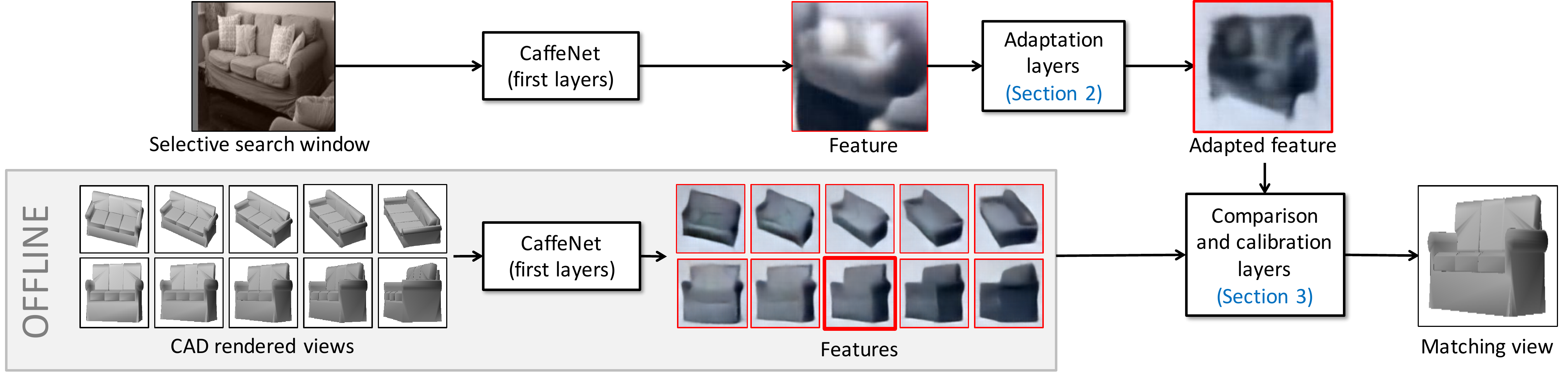}
\caption{
{\bf System overview.}
Our system takes as input individual 2D image object proposal windows (top-left) generated by the selective search algorithm~\cite{Uijlings13}.
The image window is passed through the initial layers of a pre-trained CaffeNet model~\cite{jia2014caffe} to generate a feature vector (top-middle).
Here, we visualize CNN features using the inversion network of \cite{Dosovitskiy15} \textcolor{red}{(outlined in red)}, which infers the original image given a CNN layer's response.
In an offline step (bottom-left), we similarly pass rendered views of a library of 3D object CAD models through the initial layers of CaffeNet and record their responses.
As there is a domain gap between the appearance of natural images and rendered views of CAD models, we learn to adapt the features for a natural image to better align to those of CAD models (top-right).
We compare the features and return the view that best matches the style and pose of the input image (bottom-right).
}
\label{fig:overview}
\vspace{-10pt}
\end{figure*}
Despite recent progress on 2D-3D matching and retrieval~\cite{Huang15,Li15,Su14}, detection by 2D-3D alignment lags behind state-of-the-art object detection systems based on annotated images,
e.g., R-CNN~\cite{girshick14CVPR}, in terms of accuracy and speed.
We see two primary reasons for this gap in performance:
(i) there is a large appearance gap between views rendered from CAD models and real images;
and (ii) 2D-based object detection has benefited from recent successes of convolutional neural networks (CNNs)~\cite{krizhevsky2012imagenet,Lecun}.
This work addresses both issues.

The appearance gap across two different domains encountered in 2D-3D alignment is not unique to our problem and can be found in other tasks, e.g., when learning on one dataset and testing on another~\cite{Torralba11}.
To bridge such appearance gaps, a number of cross-domain adaptation algorithms have been developed, e.g.~\cite{Tzeng15}. 
Building on the successes of these methods, 
we present an approach that learns to adapt natural image features for the task of 2D-3D exemplar detection.
We hypothesize that, given the features of a natural image depicting an object, it is possible to infer the features of a corresponding rendered view of an object CAD model with similar style and pose.
Note that similar reasoning has been explored in recent work to predict CAD object features for a different view~\cite{Guibas}.
%

To achieve our adaptation learning goal, we need a large training set of aligned natural image and rendered view pairs depicting a similar object.
While there are existing datasets with aligned pairs, e.g., IKEA~\cite{Lim13} and Pascal3D~\cite{Xiang14}, such datasets are either relatively small or have aligned models that coarsely approximates the object style.
To overcome these challenges, we make use of the ability to render views from CAD models and composite with natural images, which allows us to create a large training set.
The composite image and rendered view pairs form training data with which to learn the feature adaptation, and have been similarly employed in prior work to train 2D object detectors over CAD renders~\cite{Peng15,Pepik15} and predict object pose~\cite{Su15}.

In learning the adaptation, we adopt a formulation similar to Lenc and Vedaldi~\cite{Lenc15}, which studied the equivariance of image features under geometric deformations of the image.
Our work can be seen as an extension of their approach beyond geometric transformations.
We show that the adaptation can be incorporated as a module in a CNN-based object detection pipeline.
Furthermore, we show that pre-computed features of the rendered views can be added as a fully-connected layer in a CNN, which brings the benefits of accuracy and speed from recent advances in deep learning to 2D-3D exemplar detection.

\paragraph{Contributions.}
Our contributions are twofold:

\begin{itemize}
\item We introduce a cross-domain adaptation approach for 2D-3D exemplar detection using generated pairs of rendered views of CAD models and composite views with natural background. Our adaptation routine 
adapts features of natural images depicting objects to more closely match features of CAD model rendered views.
\item We show how our adaptation routine can be incorporated into a CNN-based detection pipeline, which leads to an increase in accuracy and speed for 2D-3D exemplar detection.
\end{itemize}

\noindent
We evaluated our method on the tasks of CAD instance retrieval on the IKEA dataset~\cite{Lim13} and on 2D-3D object class detection on the Pascal VOC subset used in Aubry \etal~\cite{Aubry14b}.
We show state-of-the-art exemplar detection performance on 
IKEA instances and out-perform the discriminative element approach of Aubry \etal~\cite{Aubry14b} both in terms of accuracy and speed.
\francisco{The extended annotations for the IKEA object dataset, a new diverse dataset of textured and non-textured rendered views of CAD models we used to learn the adaptation, and our full code are available at \url{http://imagine.enpc.fr/~suzano-f/exemplar-cnn/}.}




\subsection{Related Work}

A 3D understanding of 2D natural images has been a problem of interest in computer vision since its very beginning~\cite{Roberts65}.
Our work is in line with traditional geometry-centric approaches for object recognition based on alignment~\cite{Mundy06}.
There has been a number of successful approaches for instance-level recognition, e.g., \cite{Chum07,Li12,Rothganger06}, typically based on SIFT matching~\cite{Lowe04} with geometric constraints.
More recent approaches have leveraged contour-based representation to align skylines~\cite{Baatz12} and statues~\cite{Arandjelovic11}.
Furthermore, simplified or parametric geometric models have been used for category recognition/detection~\cite{fidler13,BlockWorld10,Hejrati12,Pepik3D,Xiao12,Zia13}.
We will focus our discussion in this section on prior work using CAD models for category recognition and 2D-3D alignment.

Rendered views from CAD models have been used as input for training an object class detector~\cite{Peng15,Pepik15,Sun14} or for viewpoint prediction~\cite{Su15}.
Most similar to us are approaches that align models directly to images.
Examples include alignment of IKEA furniture models to images~\cite{Lim13},
exemplar-based object detection~\cite{malisiewicz-iccv11} by matching discriminative elements~\cite{Aubry14b,Choy15},
and using hand-crafted features for retrieving CAD models for depth prediction~\cite{Su14} and compositing from multiple models~\cite{Huang15}.
Also related are approaches for CAD retrieval given RGB-D images (e.g., from Kinect scans)~\cite{Gupta15b,Song14}.
More recently there has been work to enrich the feature representation for matching and alignment using CNNs, which include
CAD retrieval based on CNN responses (e.g., AlexNet~\cite{krizhevsky2012imagenet} ``pool5'' features)~\cite{Aubry15},
learning a transformation from CNN features to light-field descriptors for 3D shapes~\cite{Li15},
and training a Siamese network for style retrieval~\cite{Bell15}.
Building on efficient CNN-based object class detection, e.g., R-CNN~\cite{girshick14CVPR}, our approach extends the above CNN-based approaches for efficient CAD-exemplar detection.

Bridging two very different image modalities is a classic problem for alignment~\cite{Irani98}.
Past approaches have addressed this problem using two main strategies. A first line of work has used manually-designed feature detectors and adapted them, for example by adding a mask, so that they focus on the information available in both CAD models and real images \cite{Aubry14b,Choy15,vazquez2014virtual}.
Another line of work has focused on increasing the realism of rendered views, e.g., by extracting likely textures and background from annotated images~\cite{Peng15,Pepik15,Su15,Sun14}.
Domain adaptation approaches have been formulated for CNNs~\cite{Bengio11,hinton2015distilling,romero2014fitnets,ganin2015unsupervised}, most recently for object detection~\cite{Hoffman14}, fine tuning across tasks~\cite{Tzeng15}, and, in a contemporary work, transfer learning from RGB to optical flow and depth~\cite{Gupta15}.
Most similar to our approach is domain adaptation with CAD~\cite{Sun14}, which adapted hand-crafted features (HOG~\cite{Dalal05}) for object detection.
We formulate a generic domain adaptation approach over image features, which can be applied to hand-crafted features, e.g., HOG~\cite{Dalal05} or CNN responses.

\subsection{Overview}

Figure \ref{fig:overview} shows our 2D-3D exemplar detection pipeline.
We start by computing CNN features for an image corresponding to a selective search window, along with CNN features for rendered views of CAD models.
Due to the large appearance gap across the two domains, we learn how to adapt features of natural images to better match features for rendered views (Section~\ref{sec:adaptation}).
We then compare the adapted features with calibrated rendered view features to obtain matching scores for each rendered view (Section~\ref{sec:exemplar_detection}).
Note that our detection pipeline can be implemented as a CNN.
An evaluation of our approach is in Section~\ref{sec:results}.

\section{Adapting from real to rendered views}
\label{sec:adaptation}

In this section we describe our approach for adapting features extracted from real images to better correspond to features extracted from rendered views of CAD models.
Our approach is general and can be applied to any image feature set, e.g., HOG~\cite{Dalal05} and CNN-based features~\cite{Lecun}.
We adapt from real images to rendered views (and not from rendered to real) since it is likely more difficult to hallucinate features corresponding to missing image details, such as the surrounding context of an object and its texture, than to remove them.

Formally, we seek to learn a transformation $\p$ over the features of real images.
Intuitively $\p$ is a projection of the real image feature space to the space of features from CAD rendered views.
Ideally, $\p$ has the property of mapping a given real image feature depicting an object of interest to features of rendered views of CAD object models with the same geometry, style, and pose.

Suppose we have as input a set of $N$ pairs of features $\{(x_i, y_i)\}_{i=1}^N$ corresponding to examples of real images and rendered views of well-aligned CAD models, respectively.
We seek to minimize the following cost over $\p$:

\begin{equation}
L(\p)=-\sum_{i=1}^N S\left(\p\left(x_i\right),y_i\right)+R(\p),
\label{eq:loss}
\end{equation}

\noindent
where $S$ denotes a similarity between the two features $\p(x_i)$ and $y_i$, and $R$ is a regularization function over $\p$.
Note that in the case where $\p$ is an affine transformation, our formulation is similar to the one of Lenc and Vedaldi~\cite{Lenc15} where a mapping was learned given image pairs to analyze the equivariance of CNN features under geometric transformations.


\paragraph{Adaptation.}
While the simplest choice for $\p$ is an affine transformation, which we use as a reference in our experiments, we also tested more constrained and complex transformations.
We focused on transformations that could be formulated as CNN layers, and in particular successions of convolutional and ReLU layers. Note that considering more complex transformations also increases the risk of overfitting. Similar to Lenc and Vedaldi~\cite{Lenc15} we attempted to constrain the structure of the transformation and its sparsity. This is easily done in a CNN by replacing a fully-connected layer by a convolutional layer with limited support, which implies translation invariance in the adaptation. We found that the best-performing transformation was only a slight modification of the affine transformation:

\begin{equation}
\p(x)=ReLU(Ax+b),
\label{eqn:adapt}
\end{equation}

\noindent
where $ReLU(x) = \max(0,x)$ is the element-wise maximum over zero.
We observed that applying the $ReLU$ function consistently improved results, and is in agreement with state-of-the-art CNN architecture design choices for object recognition.


\paragraph{Similarity.}
We tried both $L_2$ and squared-cosine similarity to measure the similarity in Equation~(\ref{eq:loss}).
We found that the squared-cosine similarity $S(a,b) =-\left(1-\frac{a^T b}{\|a\| \|b\|}\right)^2$ leads to better results.  
This is expected, since cosine similarity is known to work better when comparing CNN features~\cite{Aubry15}, but also because we later used the cosine distance to compare real and synthetic features (c.f.\ Section~\ref{sec:results}). This result is also consistent with the observation of the importance of task-specific similarities in Lenc and Vedaldi~\cite{Lenc15}.

\paragraph{Training data details.}
Our adaptation formulation requires a large training set of well-aligned pairs of images and rendered views of CAD models matching the style and pose of depicted objects.
Such a dataset is difficult to acquire.
While existing datasets have object CAD models aligned to images closely matching the depicted object pose~\cite{Xiang14,GuoHoiem}, the models are often not similar in style.
%
\setlength{\columnsep}{10pt}
\begin{wrapfigure}{r}{0.4\columnwidth}
  \vspace{-10pt}
  \centering
  \includegraphics[width=0.38\columnwidth]{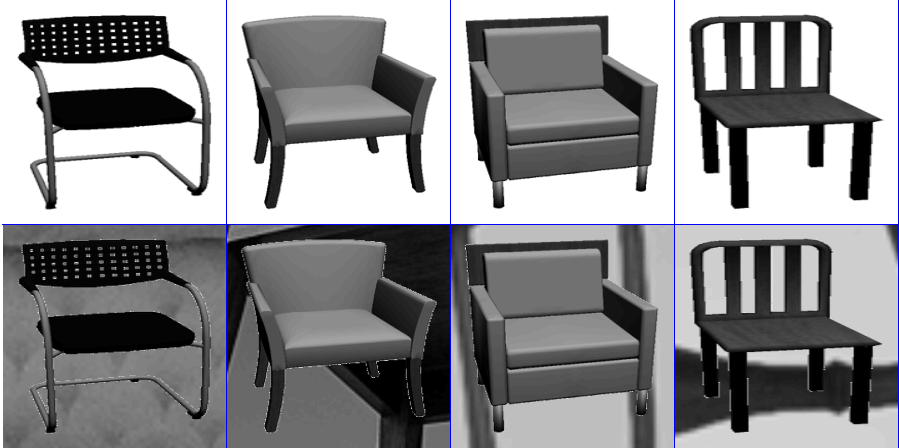}
  \caption{Examples of image pairs used for learning the adaptation.
  }
  \label{fig:data}
  \vspace{-13pt}
\end{wrapfigure}
Recent work on accurate alignment to 3D models by composition~\cite{Huang15} and semi-automatic 3-sweep modeling~\cite{CohenOrSiggraphAsia2013} are promising approaches for obtaining accurate image-model alignments, but no large-scale results are yet available.

Instead, we build on recent approaches for effective training from rendered views~\cite{Peng15,Su15} to render views of CAD models and composite on natural image backgrounds. This gives us access to virtually unlimited training data.
The backgrounds provide ``natural-looking'' surrounding context and encourages the transformation $\p$ to learn to subtract away the background context.
To avoid color artifacts in the composite images, we used gray-scale image pairs and also used gray-scale images at test time.
Note that contrary to prior approaches using manually-annotated scenes to increase the realism of the composite~\cite{Peng15,Pepik15}, we do not directly use any object annotation in our background selection process.
\francisco{Figure~\ref{fig:data} shows four representative image pairs from our adaptation data (top -- object rendered views; bottom -- rendered views composited with natural image backgrounds).}

For the 3D models, we found that using a diverse database comprising several object categories produced better results than focusing on a target set of 3D models we aim to detect. We used as reference in all our experiments the textureless rendered views from Aubry and Russell~\cite{Aubry15} to train the adaptation.

\paragraph{Implementation details.}
We used a small $L_2$ regularization $R$ in all our experiments and found that it improved our results despite our very large training sets.
We trained $\p$ using stochastic gradient descent within the Torch7 framework~\cite{collobert2011torch7}. We used a weight decay of 5e-4, corresponding to the $L_2$ regularization, a momentum 0.9, 
and mini-batch size of 128.
We started with a learning rate of 1 and reduced it every 15 epochs by a factor of 10 until convergence.

\section{Exemplar detection with CNNs}
\label{sec:exemplar_detection}

In this section we show how the adaptation procedure in Section~\ref{sec:adaptation}, together with feature computation and exemplar-based retrieval, can be incorporated into an efficient CNN-based detection routine, similar to R-CNN~\cite{girshick14CVPR}, for 2D-3D exemplar detection.
For a given input image, we seek to detect the bounding box location of an object in the image and return a corresponding CAD model having similar style, along with the pose of the depicted object.

\paragraph{Exemplar-detection pipeline.}
Following the initial part of the R-CNN object detection pipeline~\cite{girshick14CVPR}, we first extract a set of selective search windows~\cite{Uijlings13} and compute CNN responses $x$ at an intermediate layer (e.g., CaffeNet {\it pool5} layer) for each window.
We then apply our adaptation $\p$ to these features and compare the results $\p(x)$ to the features of different CAD model rendered views.
Let $s_i(x)=S(\p(x),y_i)$ be the similarity between $\p(x)$ and the features $y_i$ of the $i$th rendered view.

As shown in Aubry \etal~\cite{Aubry14b}, calibration is an important step for comparing similarity across different views and CAD models. Starting from the initial similarity score $s_i(x)$,
we apply their affine calibration routine to compute a new calibrated similarity $s'_i(x) = c_i s_i(x) + d_i$.
The scalar parameters $c_i$ and $d_i$ are selected using a large set of random patches such that $s'_i(x_0)=-1$ and $s'_i(x_1)=0$, where $x_0$ and $x_1$ correspond to random patch features with mean and 99.99-percentile similarity scores, respectively.


We take advantage of the fact that in an exemplar-based detection setup the expected aspect ratio of the alignments are known.
We remove candidate rendered-view alignments when the aspect ratio has a difference of more than $10\%$ between the selective search window and rendered view.
Finally, we rank the remaining alignments by their score $s'_i(x)$ and perform non-maximum suppression to obtain the final detections.

\paragraph{CNN implementation.}
Figure~\ref{fig:overview} shows our CNN for 2D-3D exemplar detection.
Our network starts with layers corresponding to a CNN trained on a different task (e.g., CaffeNet~\cite{jia2014caffe} trained for ImageNet classification in our experiments) until an intermediate layer (e.g.,``pool5''). 
Next, the resulting features pass through the adaptation layers corresponding to $\p$, implemented as a fully-connected layer followed by a ReLU.

The resulting adapted features are compared to the exemplar rendered-view features.
Several standard similarity functions, such as dot product and cosine similarity, can be implemented as CNN layers.
For example, cosine similarity can be implemented by a feature-normalization layer followed by a fully-connected layer. 
The weights of the fully-connected layer correspond to a matrix $Y$ of stacked unit-normalized features for the exemplar rendered views, computed in an offline stage.
While the affine calibration could be implemented as an independent layer, we incorporated it directly into the fully-connected layer by replacing the matrix rows by $Y_i \leftarrow c_i Y_i$ and adding a bias $d_i$ corresponding to each row $i$.
The final exemplar rendered-view scores is $Y \phi(x) + d$ given image features $x$, and can be computed by a single forward pass in a CNN.

\section{Experiments}
\label{sec:results}

\begin{table}
  \footnotesize
  \begin{center}
    \begin{tabular}{|c|c|c|c|c|c|c|c|c|c|}
      \hline
      & $L_2$ distance & Dot product & Cosine distance
      \\
      \hline
      Pool3 & 57.7 & 46.0 & 61.3
\\
      \hline
      Pool4 & 57.7 & 47.5 & {\bf 65.0}
            \\
      \hline
      Pool5 & 38.7 & 54.7 & 60.6
            \\
      \hline
      fc6 & 38.7 & 59.9 & 59.9
          \\
      \hline
      fc7 & 48.2 & 61.3 & 52.6
          \\
      \hline
    \end{tabular}
    \caption{
    Instance retrieval accuracy on the IKEA dataset~\cite{Lim13} over different CaffeNet layers (rows) and distances (cols).
    }
    \label{tab:ikea_retrieval_similarity}
  \end{center}
\vspace{-20pt}
\end{table}

\begin{table*}[t!]
 \small
 \setlength{\tabcolsep}{5.5pt}
  \begin{center}
    \begin{tabular}{|c||c|c|c|c|c|c|c|c|c | c|}
      \hline
      \multirow{2}{*}{class} & chair  & bookcase  & sofa  & table   & bookcase & desk & bookcase&  bed  & stool &  \\
      & \emph{poang} &\emph{billy1} &\emph{ektorp} &\emph{lack}  &\emph{billy2} & \emph{expedit} & \emph{billy4} &  \emph{malm2}  & \emph{poang} & mAP\\
      \hline
      \multicolumn{11}{c}{Original annotations of \cite{Lim13}}\\
      \hline
      Number of instances & 40 & 18 & 13 & 20  & 10 & 8 & 6 & 7  & 7 & \\
      \hline
      \hline
      Lim \etal~\cite{Lim13}  & 27.0 & {\bf 24.3} & 7.3 &  14.0 & {\bf 26.6} & 18.8 & 32.6 & {\bf 22.6} & 14.8  & {\bf20.89}\\
      \hline
      DPM \cite{lsvm-pami} & 27.5 & {\bf 24.3} & {\bf 12.1} & 10.8 & 13.5 & {\bf 46.1} & 0.1 & 1.0 & 0.5 & 15.10 \\
      \hline
      Ours without adaptation & 15.9 & 1.3 & 8.4 & 25.7 & 0.1 & 25.5 & 0.2 & 0.9 & 18.3 & 10.69\\
      \hline
    Ours with {\it fc} adaptation & 31.1 & 1.2 & 9.1 & 27.0 & 0.0 & 13.8 & 0.4 & 1.1 & 23.7 & 11.94\\
      \hline
      Ours with {\it fc+ReLU} adaptation & {\bf 33.1} & 1.1 & 9.5 & {\bf 27.1} & 0.1 & 14.5 & 0.4 & 1.2 & {\bf 24.4} & 12.38\\
      \hline
      \multicolumn{10}{c}{New annotations}\\
      \hline
      Number of instances & 56 & 35 & 15 & 34  & 17 & 7 & 17 &  24  * & 18 & \\
      \hline
      \hline
       Lim \etal~\cite{Lim13} & 19.9 & { \bf \color{red}  14.8} &{\bf 6.4} &  9.4 & {\bf \color{red} 15.7} & {\bf 15.4} & {\bf \color{red} 13.4} & {\bf 7.6} & 6.4 & { 12.11} \\
      \hline
      Ours with {\it fc+ReLU} adaptation & {\bf \color{blue} 35.4} & 6.2 & {\bf 8.2} &  {\bf \color{blue} 21.4} & 0.1 & {\bf16.5} & 0.4 & {\bf 10.4} & {\color{blue} \bf 28.4} & {\bf 14.11} \\
      \hline
    \end{tabular}
    \caption{Instance detection performance on the IKEA object dataset~\cite{Lim13}. We report average precision using a bounding box overlap threshold of 0.5. Note that some categories reported in \cite{Lim13} have very few annotated examples.  We report results for classes that include more than 3 annotated instances. 
    The top part of the table presents results with the original annotation of \cite{Lim13} and the bottom part with our extended annotations. We evaluated the detection outputs provided from \cite{Lim13} using these extended annotations.  \textsuperscript{*} The dataset includes three different but similar sizes of the same bed. Since we were not able to differentiate visually between these three kind of beds, all were annotated.
     }
    \label{tab:ikea_main}
  \end{center}
  \vspace{-15pt}
\end{table*}

In this section we qualitatively and quantitatively evaluate our method and analyze different design choices.
First, we focus on a simpler retrieval task to select the features and similarity function for our detection task (Section~\ref{sec:retrieval}).
Then, we present our main results on object-instance and object-class detection by aligning to CAD rendered views, comparing against existing baselines (Section~\ref{sec:detection}).
Finally, we perform an ablative analysis of our algorithm (Section~\ref{sec:analysis}) and report computational running time (Section~\ref{sec:time}).
\subsection{Instance retrieval}
\label{sec:retrieval}

To select CNN features and a similarity function for comparing natural images and CAD rendered views,
we consider a retrieval task where, given a cropped image depicting a query object, we seek to return a model corresponding to the object.
We consider the IKEA dataset of Lim \etal~\cite{Lim13}, which has CAD models of IKEA object instances manually aligned to their location in images depicting cluttered scenes.
The task allows us to compare the performance of different CNN layer responses and similarity functions.
The retrieval task is difficult as there are a variety of object poses and perspective effects in the IKEA dataset.
To handle the variation in object pose and perspective effects, we rendered 36 azimuth and 7 elevation angles and at 3 different distances for each object.
Note that the rendered views cover many possible viewpoints and perspective effects, but it does not cover all cases.

We extracted CNN features from CaffeNet~\cite{jia2014caffe} for our experiments.
While more recent, deeper networks~\cite{Simonyan14c,szegedy2014going} may yield better results
(e.g., a boost of $4\%$ is obtained for retrieval using the cosine distance on VGG {\it pool4} features), we illustrate the basic design choices using the shallower CaffeNet model.
We also expect better results using the last layers of a network fine tuned for object-class detection, e.g., R-CNN fine tuned for Pascal detection~\cite{girshick14CVPR}.
We chose not to consider such a network to focus on the general case where natural images of related object classes do not have to be annotated for training.
We performed retrieval using features extracted from the {\it conv3} to {\it fc7} layers of CaffeNet after ReLU (and without adaptation).
We applied max-pooling to the {\it conv3} and {\it conv4} features to keep their dimensionality relatively small and avoid memory issues in our detection pipeline.
We denote the resulting features after pooling as {\it pool3} and {\it pool4}. 
We compared three similarity functions for our experiments: $L_2$ distance, dot-product similarity, and cosine distance.

We report retrieval accuracy in Table~\ref{tab:ikea_retrieval_similarity}.
Notice that performance for cosine distance is best with {\it pool4} features, and decreases with the higher layers,
while performance increases with the higher layers for dot-product similarity. 
Based on these results, we used cosine distance over {\it pool4} features in all our experiments.
Moreover, {\it conv4} features are known to be relatively generic features~\cite{pulkit,YosinskiNips14} and make little to no use of the network knowledge gained on specific objects, such as chairs, sofas, and beds, in ImageNet classification.

\subsection{Detection}
\label{sec:detection}

In this section, we demonstrate our feature-adaptation algorithm for 2D-3D detection.
We consider two tasks: object-instance and object-category detection by 2D-3D alignment.
For object-instance detection, we evaluated on the IKEA dataset~\cite{Lim13}.
For object-category detection, we evaluated on the subset of Pascal VOC containing ``chairs'' used in Aubry \etal~\cite{Aubry14b}.
We show qualitative and quantitative results on both benchmarks and compare against prior work.



\subsubsection{Object-instance detection by 2D-3D alignment}

For object-instance detection by 2D-3D alignment, we evaluated our approach on the IKEA dataset and followed the detection protocol outlined in Lim \etal~\cite{Lim13}.
We report average precision detection performance in Table~\ref{tab:ikea_main}(top), along with baselines for this task.
It can be seen that we clearly improve over the baselines for several well-represented classes.
However, our mAP is smaller than the baselines.
We will show that this is due to two main effects: a chance factor for classes where very few objects were annotated or had missing annotations, and a failure of our algorithm on ``bookcases'', which we analyze in detail.

 \paragraph{Dataset and additional annotations.}
 Two important issues when using the IKEA object dataset for evaluating instance detection are (i) its relatively small size (we report the number of annotated instances in the first line of table~\ref{tab:ikea_main}), and (ii) the partial annotations made available, with a maximum of one object per image when several are often present.
 To partly address these issues, we annotated all instances in the 288 test images for the classes that included more than three instances in the original dataset (except for ``Billy3'', where the detections reported in \cite{Lim13} appear to correspond to a different model). This increases the number of annotated objects of the selected classes from 129 to 223. We report our results on our new extended annotation set in Table~\ref{tab:ikea_main}(bottom). We will release these new annotation to allow further comparisons. With these extended annotations our mAP is similar to \cite{Lim13}, but with strong differences in the performance for the different objects. We have similar results or clear improvements (shown in blue in table~\ref{tab:ikea_main}) for most classes, but much lower performance for bookcases (shown in red in table~\ref{tab:ikea_main}).

\begin{figure}
\center
\includegraphics[width=0.19\columnwidth]{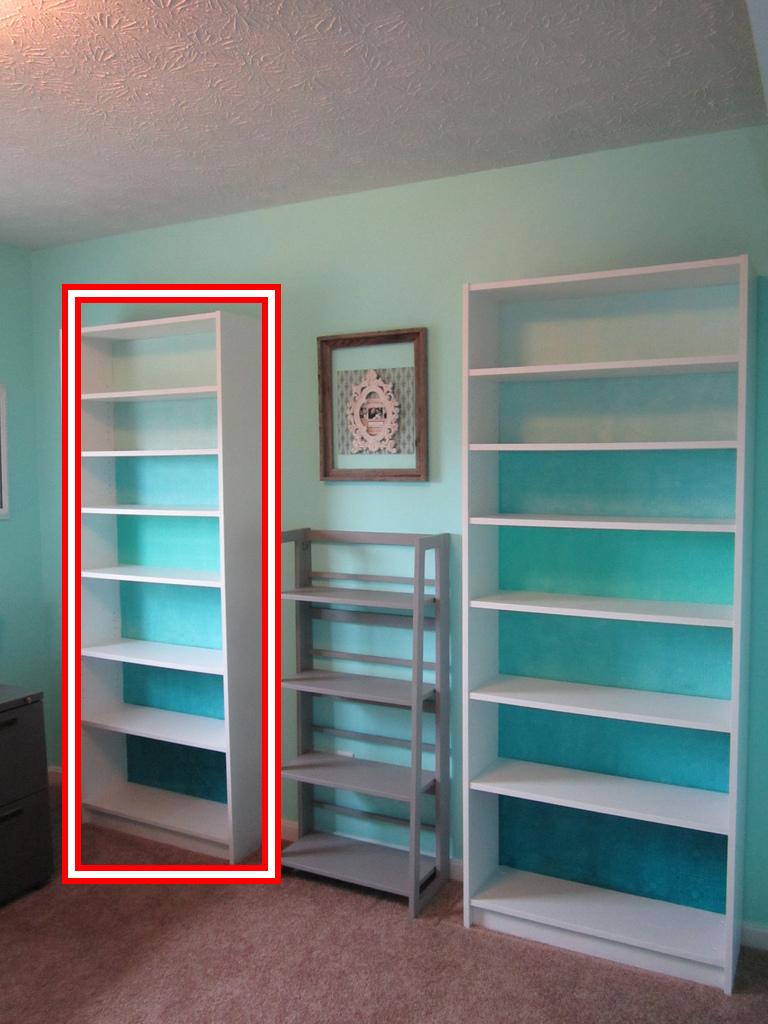}
\includegraphics[width=0.19\columnwidth]{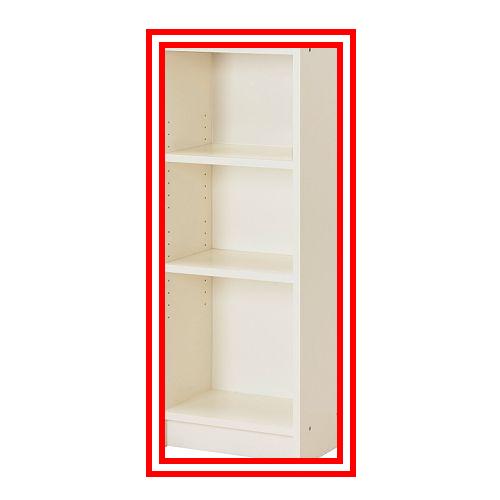}
\includegraphics[width=0.19\columnwidth]{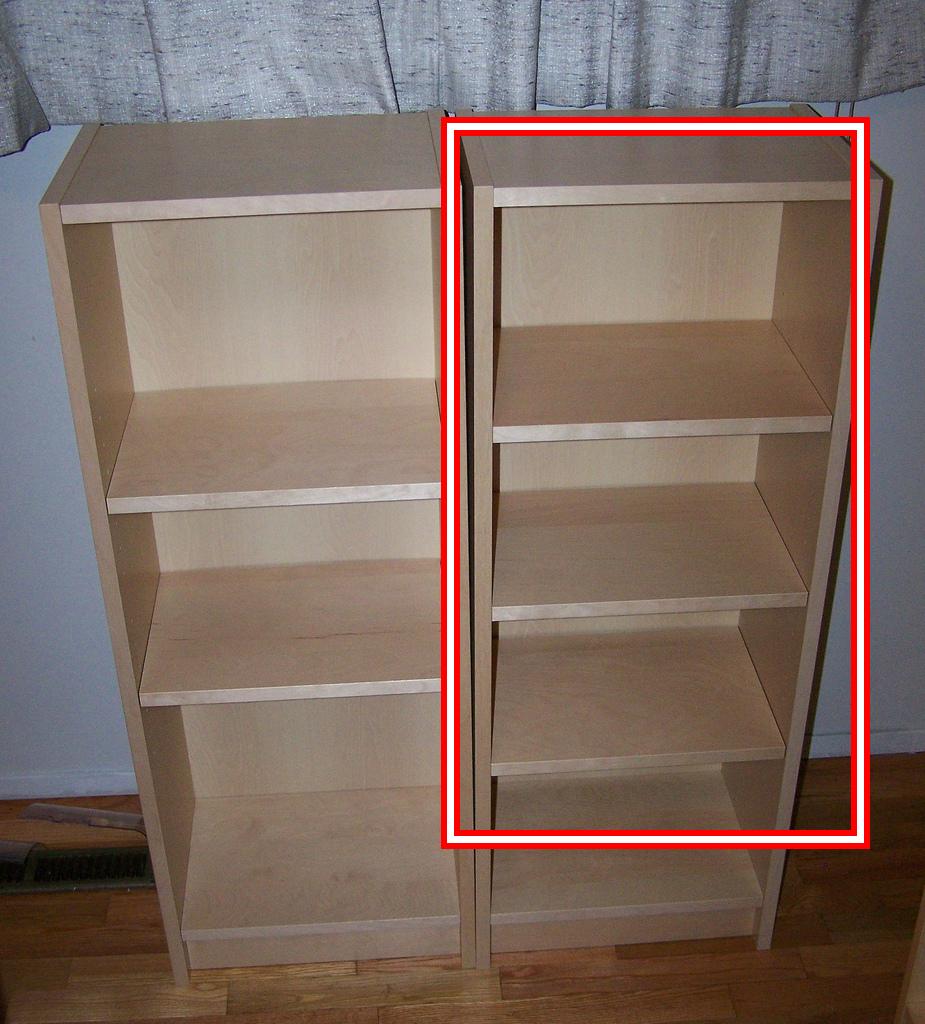}
\includegraphics[width=0.19\columnwidth]{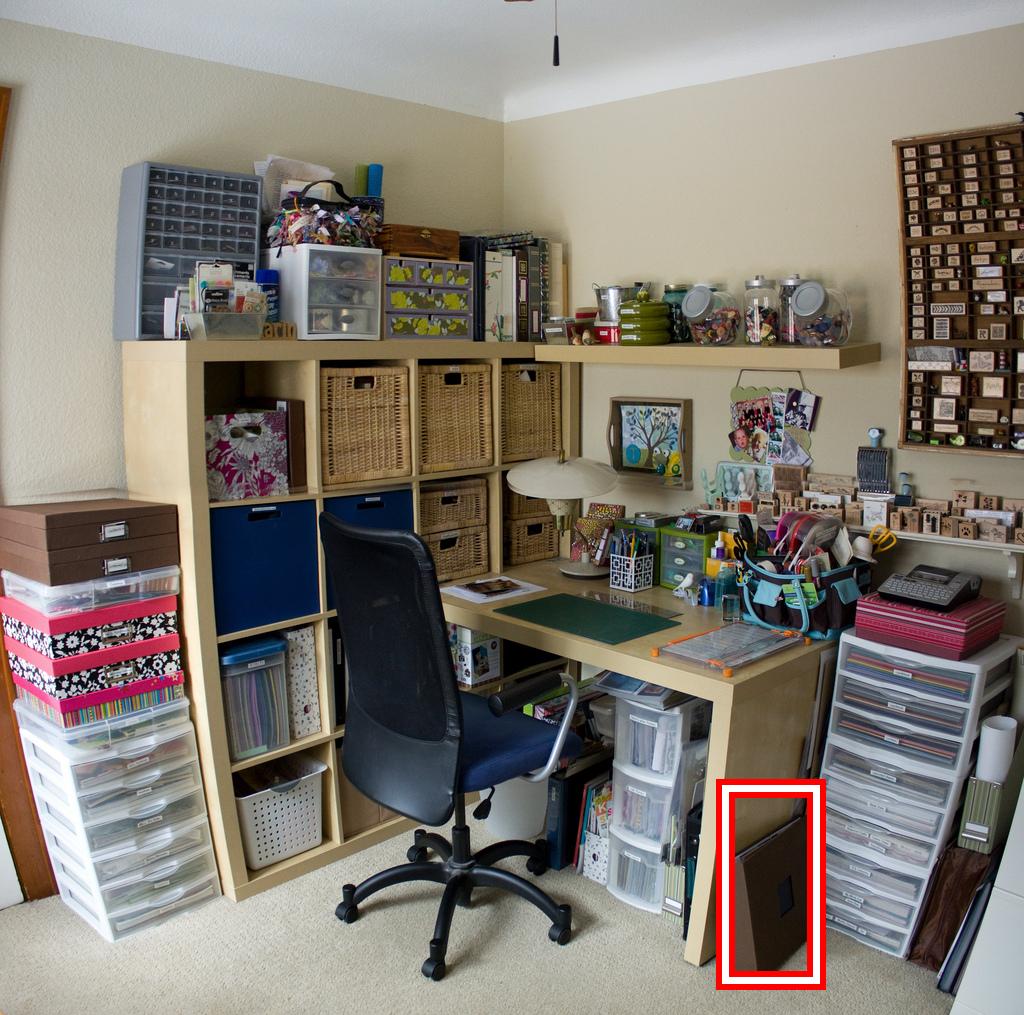}
\includegraphics[width=0.19\columnwidth]{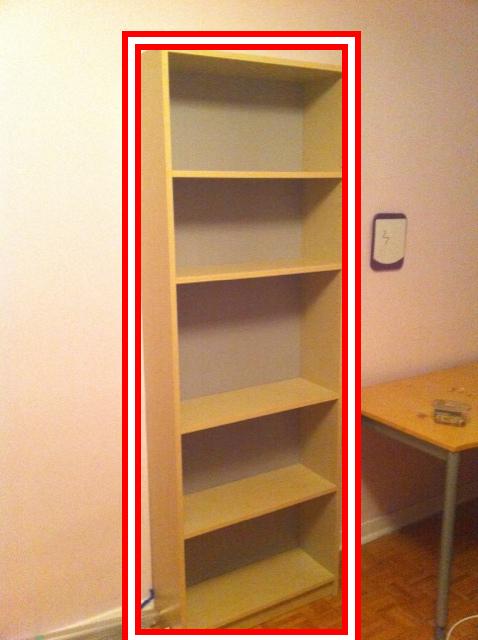}
\includegraphics[width=0.19\columnwidth]{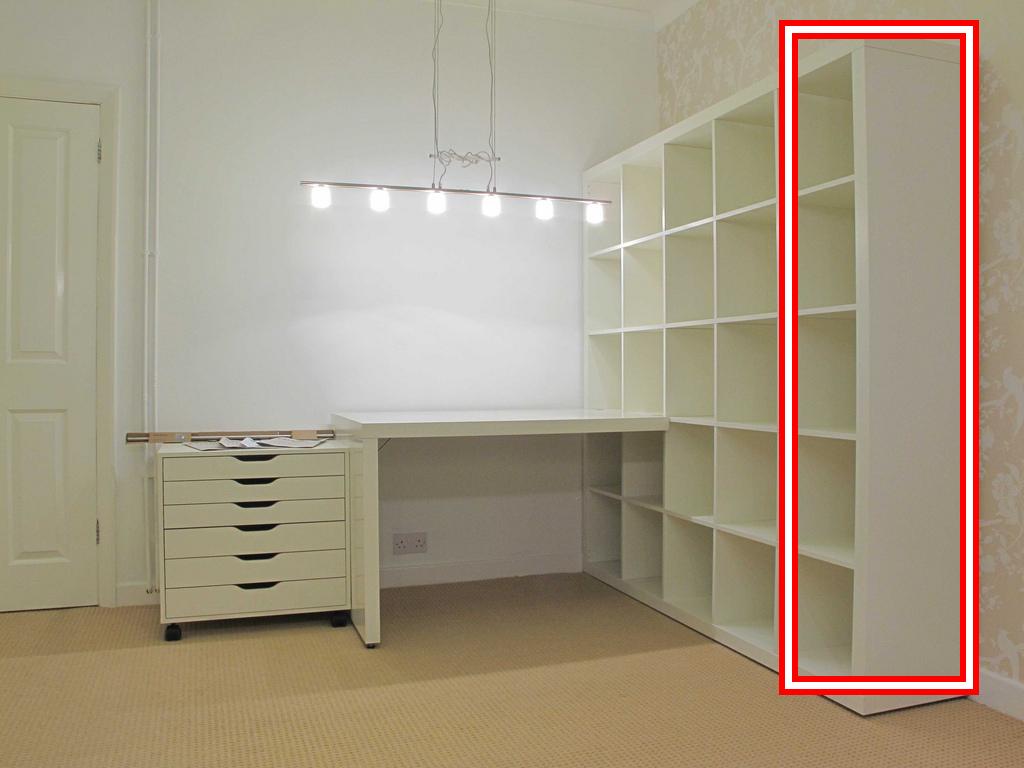}
\includegraphics[width=0.19\columnwidth]{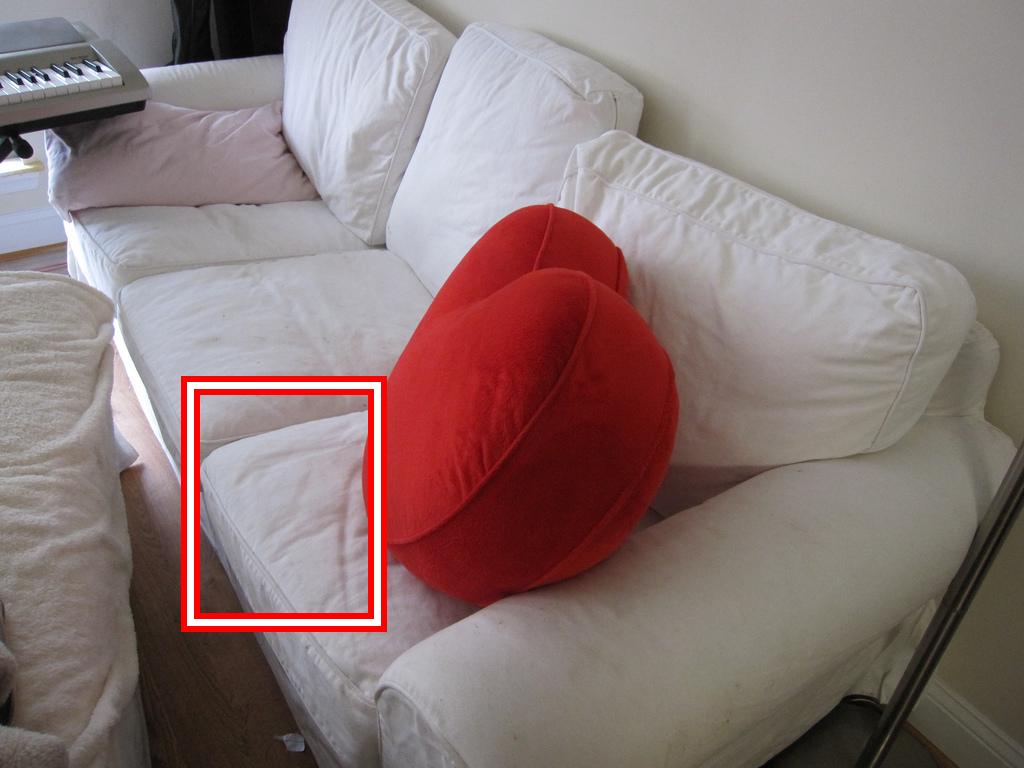}
\includegraphics[width=0.19\columnwidth]{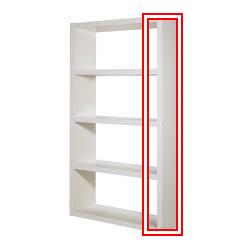}
\includegraphics[width=0.19\columnwidth]{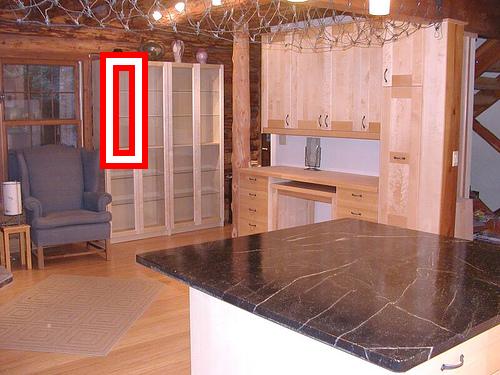}
\includegraphics[width=0.19\columnwidth]{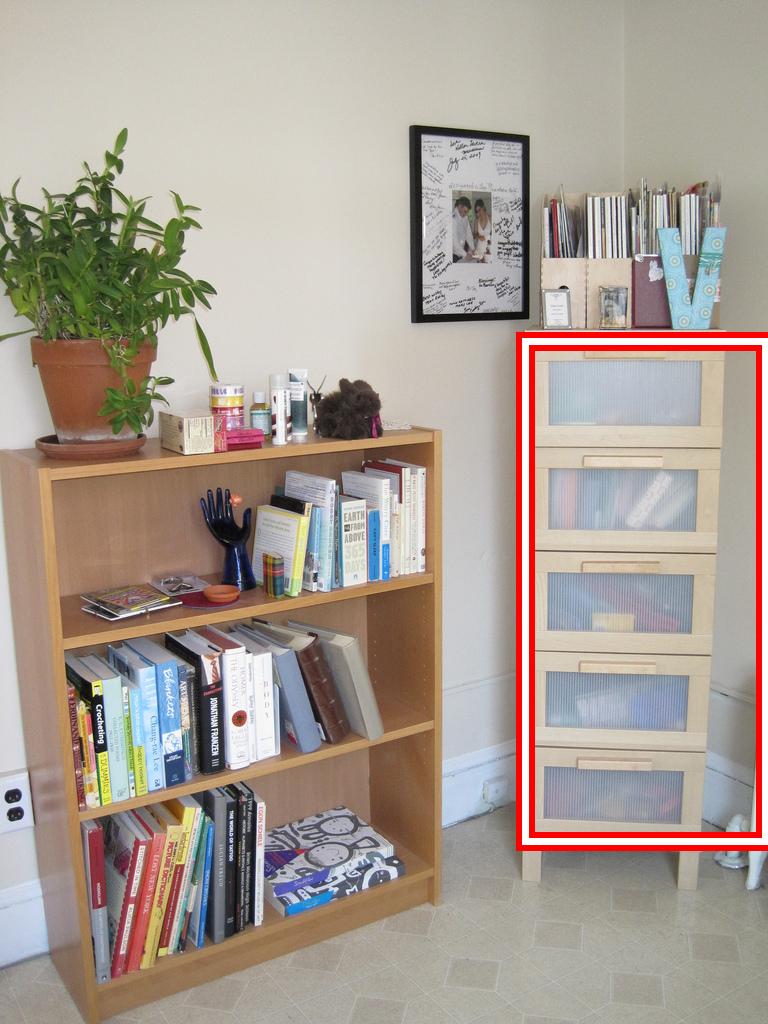}
\caption{Top 10 detections for the {\it billy1} IKEA model. Note that the first good detection is counted as negative with the original annotation because it was not annotated in the dataset. Most of our other detections are different bookcases or parts of bookcases.}
\label{fig:negbilly}
\vspace{-15pt}
\end{figure}

\paragraph{Failures on bookcases.}
Here we analyze our failures for bookcases, which are very poor in contrast to other categories where they matched or exceeded the baselines.
\francisco{Inspecting the bookcases missed by our algorithm, which are available in the project webpage, almost all of them consist of highly cluttered examples, e.g., bookcases filled with books of different colors. We verified that for our extended annotations, only $14\%$ of \emph{billy1} bookcases are empty, whereas \emph{billy2} and \emph{billy4} do not have any non-cluttered examples in the dataset.}
Looking at our top false positives in Figure~\ref{fig:negbilly} confirms this,
since we find many parts of \francisco{empty} bookcases or bookcases from other categories.



\subsubsection{Object-category detection by 2D-3D alignment}

\begin{table}
  \small
  \begin{center}
    \begin{tabular}{|c|c|c|c|}
      \hline
      \multicolumn{4}{|c|}{Training with real data} \\
      \hline
      \hline
DPM \cite{lsvm-pami} & \multicolumn{3}{|c|}{41.0} \\
      \hline
R-CNN \cite{girshick14CVPR} & \multicolumn{3}{|c|}{44.8} \\
      \hline
R-CNN + SVM \cite{girshick14CVPR} & \multicolumn{3}{|c|}{54.5} \\
      \hline
      \hline
      \multicolumn{4}{|c|}{Training with CAD data} \\
      \hline
      \hline
Aubry \etal~\cite{Aubry14b} & \multicolumn{3}{|c|}{33.9} \\
            \hline
Peng \etal~\cite{Peng15} (W-UG) & \multicolumn{3}{|c|}{29.6} \\
      \hline
      \hline
\multirow{2}{*}{} & \multirow{2}{*}{Adaptation}  & \multicolumn{2}{|c|}{No Adaptation}  \\
  \cline{3-4}
  & & Comp. & White\\
  \hline
  \hline
Logistic {\it pool4} &  12.9 & 3.7 & 1.4 \\
      \hline
Logistic {\it fc7} & 26.6 & 9.2 & 14.0 \\
      \hline
Ours, no calibration &  5.6 & 6.0 & 3.2\\
      \hline
Ours with calibration & 52.3 & 36.4 & 17.9\\
      \hline

    \end{tabular}
    \caption{ Average precision for chair detection on Pascal VOC subset \cite{Aubry14b}. Our best method outperforms the baselines of \cite{Aubry14b} by $18\%$. \francisco{``White'' column corresponds to synthetic images on white background. ``Comp'' column corresponds to synthetic images composited on real-image backgrounds.}}
    \label{tab:chair_main}
  \end{center}
\vspace{-20pt}
\end{table}

For object-category detection by 2D-3D alignment, we evaluated our approach on the subset of the Pascal VOC dataset containing images of non-difficult, non-occluded, and non-truncated ``chairs'' used in Aubry \etal~\cite{Aubry14b}, and aligned to their chair rendered views.
We followed their detection protocol and report average precision for the detection task.
We compare our performance against the baseline of Aubry \etal~\cite{Aubry14b}, which also performs detection by 2D-3D alignment.
We also report performance of DPM~\cite{lsvm-pami} and R-CNN~\cite{girshick14CVPR} \francisco{with and without SVM, both} without bounding box regression, which were trained on natural images for 2D object detection.
As another baseline,
we report the performance of a logistic regression classifier trained using synthetic images (with and without adaptation),
which is similar in spirit to recent approaches that trains a 2D object detector using synthetic training images~\cite{Peng15,Pepik15}.
\francisco{In order to better situate our work with respect to approaches that train a classifier using synthetic images with composite backgrounds~\cite{Peng15,Pepik15}, we also report results for the following baselines using synthetic images composited with natural-image background as positives, and without adaptation: (a) logistic regression classifier, (b) our exemplar detector. Finally, we report results for the best performing method of Peng~\etal~\cite{Peng15}, corresponding to their W-UG synthetic images.}

We report our results in Table~\ref{tab:chair_main}.
With our reference adaptation, our method outperforms all baselines except R-CNN + SVM. 
 We obtain an average precision of $52.3\%$ compared to $41\%$ for DPM, $33.9\%$ for Aubry \etal~\cite{Aubry14b} \francisco{and $29.6\%$ for  Peng~\etal~\cite{Peng15}}. \francisco{We also tried using the method of \cite{Peng15} with the chairs from \cite{Aubry14b}, which resulted in 9.0 AP. This difference in performance is likely due to their manual selection of realistic viewpoints and models in the W-UG set.}



  \begin{figure}
  \center
  \begin{subfigure}[b]{0.4\linewidth}
  \includegraphics[height=2cm]{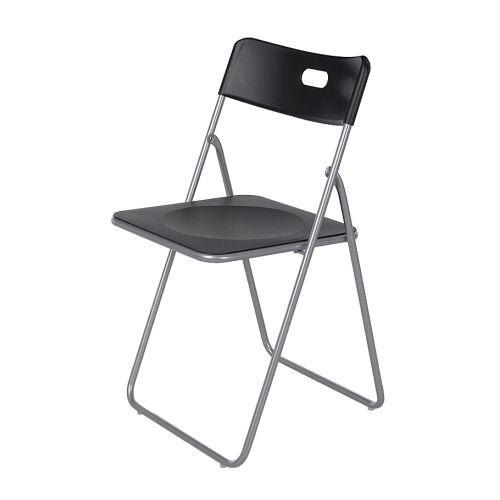}
  \includegraphics[height=2cm]{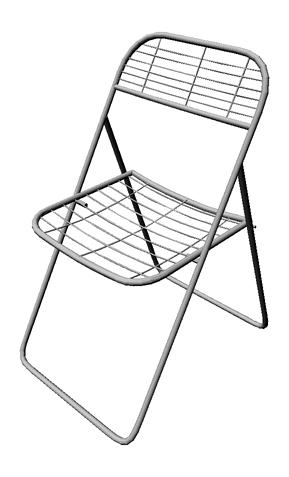}
  \includegraphics[height=2cm]{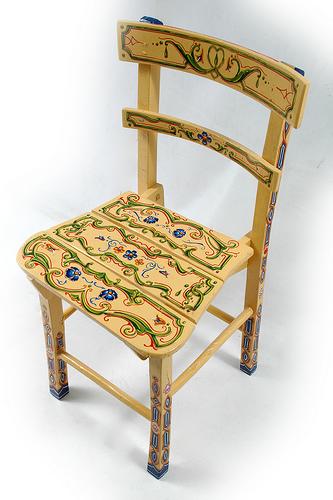}
  \includegraphics[height=2cm]{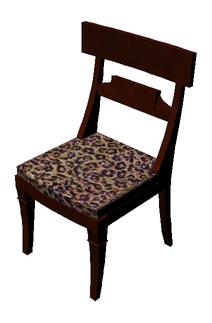}
  \includegraphics[height=2cm]{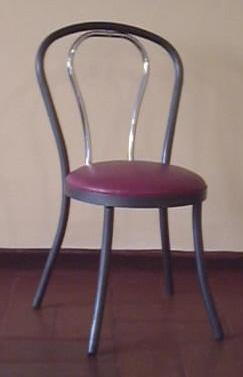}
  \includegraphics[height=2cm]{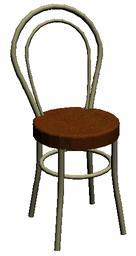}
  \caption{Without adaptation}
  \end{subfigure}
  \hspace{1mm}
  \vline
  \hspace{2mm}
  \begin{subfigure}[b]{0.4\linewidth}
  \includegraphics[height=2cm]{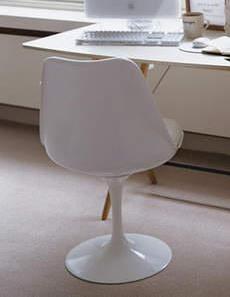}
  \includegraphics[height=2cm]{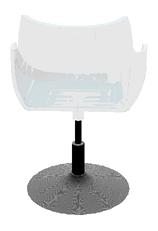}
  \includegraphics[height=2cm]{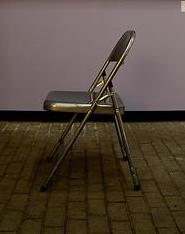}
  \includegraphics[height=2cm]{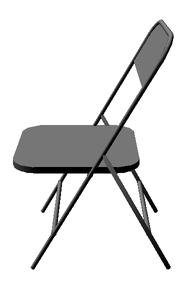}
  \includegraphics[height=2cm]{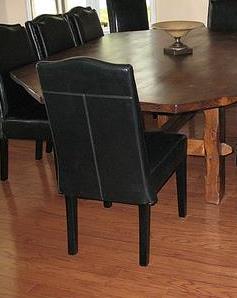}
  \includegraphics[height=2cm]{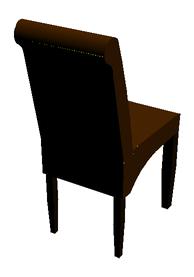}
  \caption{With adaptation}
  \end{subfigure}
  \caption{
  Top detections without and with adaptation on the Pascal VOC chair subset~\cite{Aubry14b}.
  Notice that while the alignments are good with and without adaptation, detection without adaptation returns dark chairs having ``CAD-like'' white backgrounds.
  Detections with adaptation include brighter objects and cluttered backgrounds.  }
  \label{fig:topvoc}
  \vspace{-15pt}
  \end{figure}

  A more detailed analysis reveals the importance of the adaptation for all the methods based only on CNN features from CAD models. Note that the benefit of using the adaptation is less important when using the fc7 layer for logistic regression. This shows that unsurprisingly {\it fc7} is less sensitive to the type of representation than {\it conv4}, and may explain the good results obtained by \cite{Peng15,Pepik15} using the fc7 layers directly.
\francisco{An interesting question is whether the adaptation could be replaced by  synthetic images composited with natural-image backgrounds. As can be seen from Table~\ref{tab:chair_main}, even though the composites help in some cases (notably in our exemplar detector), its performance still lags behind the performance obtained using the adaptation. Note that we used a single background per exemplar view. While one could include more composites per exemplar, this would increase the memory requirements as one would need to store all of the additional exemplars.}

\subsection{Ablative analysis}
\label{sec:analysis}

In this section we perform an ablative study of different design choices of our approach.

\begin{figure}
\center
\begin{subfigure}[b]{\linewidth}
\center
\includegraphics[height=0.18\columnwidth]{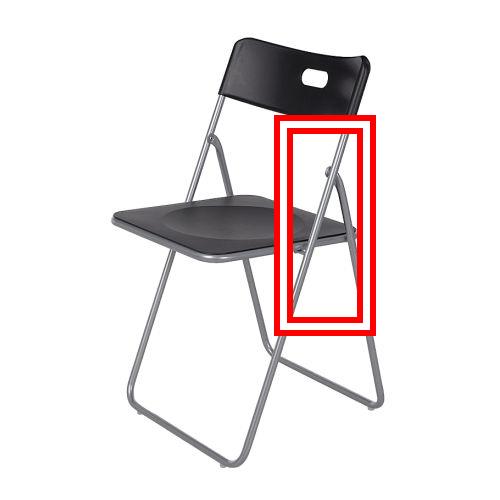}
\includegraphics[height=0.18\columnwidth]{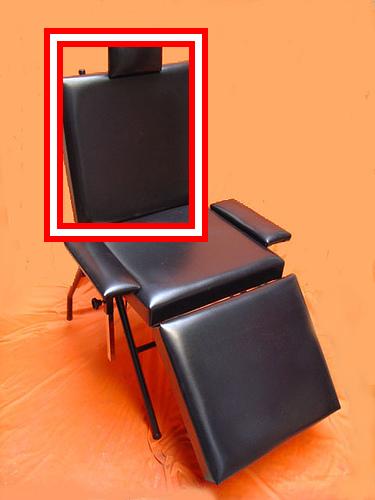}
\includegraphics[height=0.18\columnwidth]{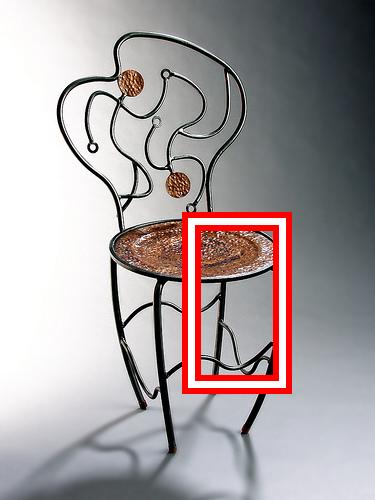}
\includegraphics[height=0.18\columnwidth]{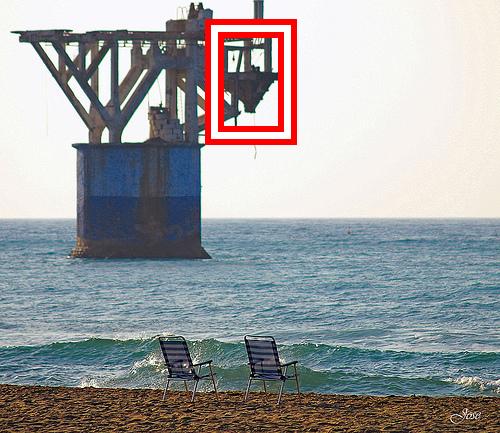}
\includegraphics[height=0.18\columnwidth]{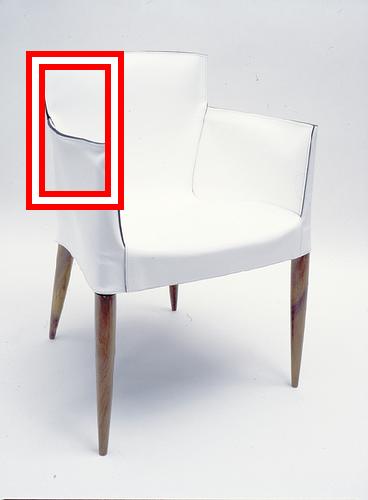}
\caption{Top false positives without adaptation}
\end{subfigure}
\begin{subfigure}[b]{\linewidth}
\center
\includegraphics[height=0.16\columnwidth]{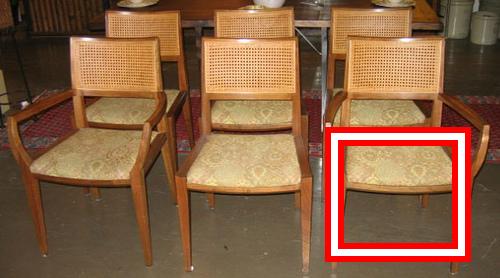}
\includegraphics[height=0.16\columnwidth]{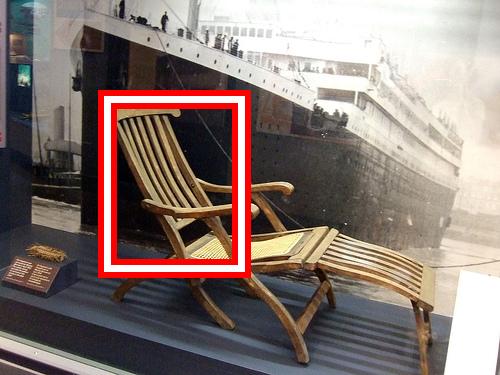}
\includegraphics[height=0.16\columnwidth]{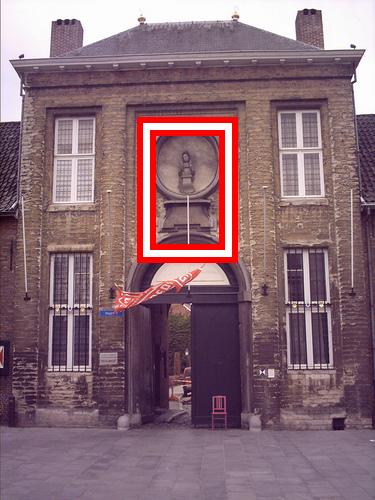}
\includegraphics[height=0.16\columnwidth]{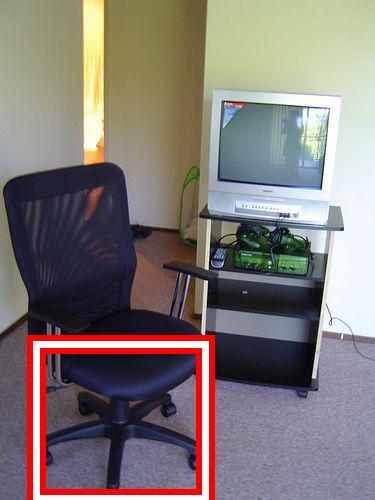}
\includegraphics[height=0.16\columnwidth]{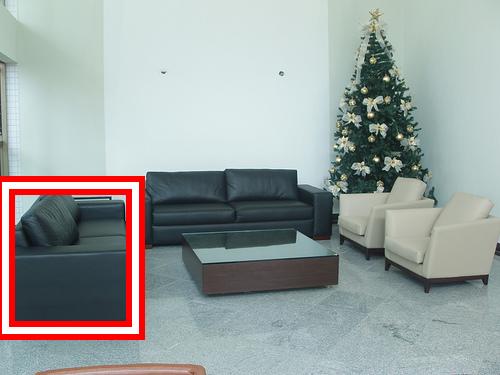}
\caption{Top false positives with adaptation}
\end{subfigure}
\caption{Top-ranked false positives without and with adaptation.
Since there were several false positives per image without adaptation, we only show the best ranked for each image.
The false positives without adaptation occur on uniform background patches.
With adaptation, this effect largely disappears and the false positives correspond to patches that look like chairs or chair parts.  }
\label{fig:negvoc}
\vspace{-3mm}
\end{figure}

\paragraph{Influence of adaptation on alignment.}
In Figure~\ref{fig:topvoc}, we show the top detections with and without adaptation.
Notice that while the non-adapted features have higher detection scores for ``CAD-like'' images of darker chairs on mostly white background (Fig.~\ref{fig:topvoc}(a)), the adaptation allows us to detect chairs of all colors in natural cluttered scenes (Fig.~\ref{fig:topvoc}(b)).
Similarly, we show the top false positives in Figure~\ref{fig:negvoc}.
Notice that without adaptation the top false positives correspond to regions with uniform background (Fig.~\ref{fig:negvoc}(a)),
while adaptation has chair-shaped false positives similar to an object detector trained on natural images only (Fig.~\ref{fig:negvoc}(b)).


\paragraph{Adaptation design.}
As discussed in Section~\ref{sec:exemplar_detection}, the adaptation $\p$ in Equation~(\ref{eqn:adapt}) can be implemented in a CNN as a fully-connected layer, followed by a ReLU nonlinearity.
We seek to study variants of $\p$.
Since the {\it pool4} CaffeNet features maintain spatial bin structure, we consider adaptations with limited spatial support via convolution with $1\times 1$ and $3\times 3$ kernels.
We also consider whether to use the ReLU nonlinearity and whether to consider multiple convolutional layers in the adaptation.

\begin{figure}
\centering
\begin{subfigure}[b]{0.5\linewidth}
\includegraphics[width=\columnwidth]{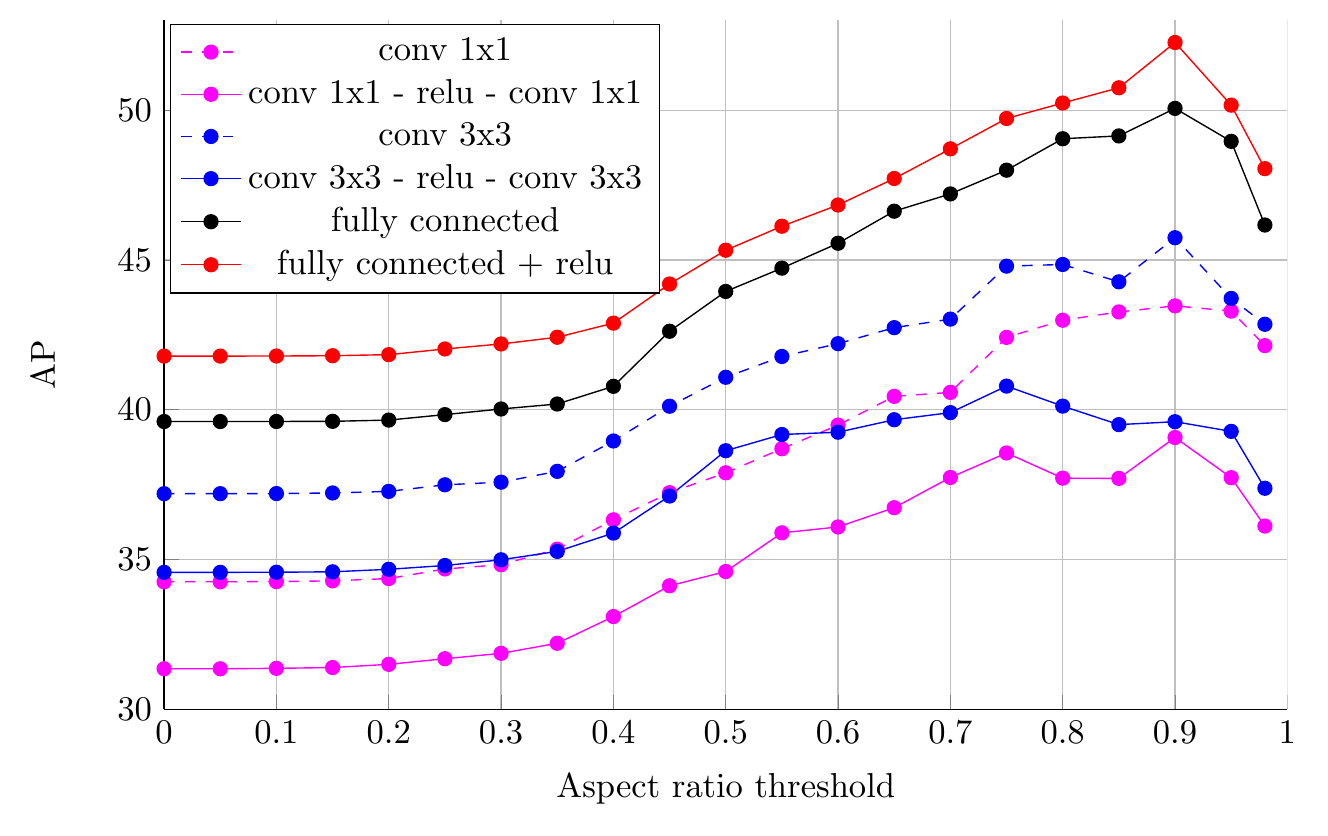}
\caption{}
\label{fig:ap_curves}
\end{subfigure}
\begin{subfigure}[b]{0.48\linewidth}
\includegraphics[width=\columnwidth]{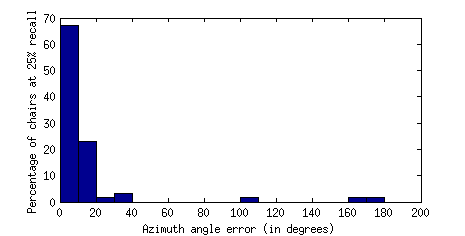}
\caption{}
\label{fig:pose}
\end{subfigure}
\vspace{-3mm}
\caption{(a) Average precision for different adaptations as a function of the aspect ratio threshold. (b) Azimuth angle error. Best viewed in the electronic version.}
\vspace{-5mm}
\end{figure}

Figure~\ref{fig:ap_curves} shows the average precision for different variants of $\p$ as a function of the aspect ratio threshold.
Notice that all of the adaptation variants we tried performed better than without adaptation ($17.9\%$ AP).
Imposing adaptations with limited spatial support ({\it conv}) performed worse than a fully-connected layer.
This can be understood by considering that the effect of the projection depends on the interpretation of the image as foreground object and background as clutter, a task that can be  better performed globally.
Using two layers for the adaptation degraded performance.
Note that we observed the validation loss was better optimized using two layers.
We believe this effect is due to the synthetic nature of our training data, which only approximates the relation between real and synthetic images.
Finally, we found that adding a ReLU after the convolutional layer consistently increased the performance.
The use of a single fully-connected layer followed by a {ReLU} produced the best performance. 

\paragraph{Aspect ratio.}
Figure~\ref{fig:ap_curves} shows the evolution of the average precision as a function of aspect ratio threshold for different projections on the Pascal VOC subset detection experiment.
As expected, increasing the threshold first improves the results because it removes many false positives.
The results are then relatively stable between 0.75 and 0.9 since both positives and negatives are discarded.
Finally, the performance drops for higher thresholds as more true positives get discarded.
In all our experiments, we used an aspect-ratio threshold of 0.90.

\paragraph{Evaluation of the retrieved pose.}
We conducted the same experiment as in Aubry~\etal~\cite{Aubry14b} to evaluate the quality of the retrieved poses.
For ground truth we used the pose annotations from Pascal3D~\cite{Xiang14}.
Figure~\ref{fig:pose} shows a histogram of azimuth angle errors at 25\% recall (similar to Fig.~6 in Aubry~\etal~\cite{Aubry14b}). Our algorithm returns an azimuth angle within $20^\circ$ of the ground truth for 90\% of the examples, compared with 87\% for Aubry~\etal~\cite{Aubry14b}.

\paragraph{Number of rendered views.}
We studied the relative importance of the CAD model dataset size on the final detection performance by conducting experiments over the set of 86K renders from Aubry \etal~\cite{Aubry14b}.
We randomly selected increasing subsets of all rendered views (Table~\ref{tab:numberstudy}(a)), and randomly selected increasing numbers of CAD models and used all their 62 rendered views (Table~\ref{tab:numberstudy}(b)). 
Notice that performance increases with the number of CAD renders\francisco{, as expected.} 
Interestingly, the diversity of the CAD models plays an important role in the final detection score.
For roughly the same number of rendered views, 5 CAD models (for a total of 310 views) performs considerably worse than 200 random views.

\subsection{Computational run time}
\label{sec:time}
Our system runs in computational time similar to R-CNN~\cite{girshick14CVPR} if all the CAD rendered views fit into GPU memory.
Excluding the time to compute bounding box proposals, we can align a test image to 2K rendered views in approximately 9.5 seconds on a GeForce GTX980 graphics card.
We can align to more views at the expense of copying pre-computed rendered view features to the GPU memory.
This can be overcome with larger-memory graphics cards or by running on parallel cards.
For 80K rendered views, our approach takes around 52 seconds.
Similar to recent fast CNN detection pipelines~\cite{he2014spatial,girshickICCV15fastrcnn}, our timings could be further optimized by reusing the convolutional features for each bounding box, which could potentially reduce the computational time to a fraction of a second.
Filtering by aspect ratio before comparing the features could also reduce the number of tests to perform, especially in the case of very large number of 3D views.
Note that even without these improvements, our computational run times are much faster than those presented in Aubry \etal~\cite{Aubry14b}.

\section{Conclusion}
\label{sec:conclusion}

We demonstrated an end-to-end CNN for 2D-3D exemplar detection.
We showed that an adaptation of image features to closely match features of rendered views of CAD models is essential to its success.
Our adaptation approach is agnostic to the feature set and could potentially benefit other 2D-3D detection methods.

\begin{table}
  \small
  \center
  \begin{tabular}{|c||c|c|c|c|c|c|}
    \hline
    \multicolumn{7}{|c|}{(a) Number of rendered views} \\
    \hline
    \hline
    $n$ & 200 & 500 & 1k & 2k  & 10k & 86k\\
    \hline
    AP & 33.3 & 37.6 & 41.3 & 44.8  & 45.7 & 50.0\\
    \hline
    \multicolumn{7}{|c|}{(b) Number of CAD models} \\
    \hline
    \hline
    $n$ & 5 & 10 & 20 & 40  & 160 & 1393\\
    \hline
    AP & 21.7 & 26.6 & 29.8 & 33.9 & 44.6 & 50.0\\
    \hline
  \end{tabular}
  \caption{Detection AP in the subset of Pascal VOC chair subset~\cite{Aubry14b} for the fully-connected projection as a function of (a) the number of CAD rendered views and (b) the number of unique CAD models used.}
  \label{tab:numberstudy}
\end{table}

\section{Aknowledgments}
We thank Joseph Lim, who shared with us his IKEA detection outputs, which allowed us to compare against his approach using our extended annotations.
We also wish to thank Alyosha Efros and Renaud Marlet for fruitful discussions.
This work was partly supported by ANR project Semapolis ANR-13-CORD-0003, Intel, a gift from Adobe, and hardware donation from Nvidia.

{\small
\bibliographystyle{ieee}
\bibliography{refs}
}

\end{document}